\documentclass[conference]{IEEEtran}
\IEEEoverridecommandlockouts
\usepackage{cite}
\usepackage{amsmath,amssymb,amsfonts}
\usepackage{mathtools}
\usepackage{algpseudocode, algorithm}
\usepackage[algo2e]{algorithm2e}
\usepackage{graphicx}
\usepackage{textcomp}
\usepackage{xcolor}
\usepackage{subfigure}
\usepackage{subfiles}

\newcommand{\rvp}{RVP }
\newcommand{\rvpfull}{Reshaping Viscoelastic-String Path-Planner }

\def\BibTeX{{\rm B\kern-.05em{\sc i\kern-.025em b}\kern-.08em
    T\kern-.1667em\lower.7ex\hbox{E}\kern-.125emX}}
\begin{document}

\title{Reshaping Viscoelastic-String Path-Planner (RVP)\\
\thanks{This work was supported in part by the Army Research Laboratories under Grant W911NF1920243.}
}

\author{ \IEEEauthorblockN{Sarvesh Mayilvahanan}
\IEEEauthorblockA{\textit{Department of Robotics} \\
\textit{University of Michigan}\\
Ann Arbor, MI, United States \\
smayil@umich.edu}
\and
\IEEEauthorblockN{Akshay Sarvesh}
\IEEEauthorblockA{\textit{ Electrical and Computer Engineering} \\
\textit{Texas A\&M University}\\
College Station, TX, United States \\
sarvesh@tamu.edu}
\and
\IEEEauthorblockN{Swaminathan Gopalswamy}
\IEEEauthorblockA{\textit{ Mechanical Engineering} \\
\textit{Texas A\&M University}\\
College Station, TX, United States \\
sgopalswamy@tamu.edu}
}

\maketitle

\begin{abstract}
We present \rvpfull(RVP) a Path Planner that reshapes a desired Global Plan for a Robotic Vehicle based on sensor observations of the Environment. We model the path to be a viscoelastic string with shape preserving tendencies, approximated by a connected series of Springs, Masses, and Dampers. The resultant path is then reshaped according to the forces emanating from the obstacles until an equilibrium is reached. The reshaped path remains close in shape to the original path because of \emph{Anchor Points} that connect to the discrete masses through springs. The final path is the resultant equilibrium configuration of the Spring-Mass-Damper network. Two key concepts enable \rvp (i) \emph{Virtual Obstacle Forces} that push the Spring-Mass-Damper system away from the original path and (ii) \emph{Anchor points} in conjunction with the Spring-Mass-Damper network that attempts to retain the path shape. We demonstrate the results in simulation and compare it's performance with an existing Reshaping Local Planner that also takes a Global Plan and reshapes it according to sensor based observations of the environment. 

\end{abstract}

\begin{IEEEkeywords}
 Path Planning, Motion Planning, Dynamic Control, Autonomous Navigation, Robotic Navigation
 \end{IEEEkeywords}

\section{Introduction}
\emph{Autonomous Mobility of Robotic Vehicles (RVs)} is a research problem that makes it's presence in a wide variety of applications. These applications include but are not limited to: Navigation applications like self-driving cars, Autonomous Trucking and Warehouse management. There is also a push to reduce the presence of humans in certain hazardous environments due to the dangers posed to their health and well-being. Some of these applications include using Robots to extinguish wildfires, Autonomous Mining, and rescue operations in disaster prone areas \cite{murphy2014disaster}. Oftentimes, due to the nature of the environment, lack of structure in the static and dynamic obstacles, the problem of \emph{Autonomous Navigation of Robots} where a \emph{RV} starts from some point in an environment and is given a mission to navigate to an end point remains a difficult one to solve. Some other applications of this are in the domain of Robot Manipulators whose applications vary widely from Robotic Surgery in the medical domain to folding clothes in the retail domain. These applications all require precise navigation of Robots in desired configurations while avoiding obstacles. 

\begin{figure}[thbp]
    \centerline{\includegraphics[width=0.9\columnwidth]{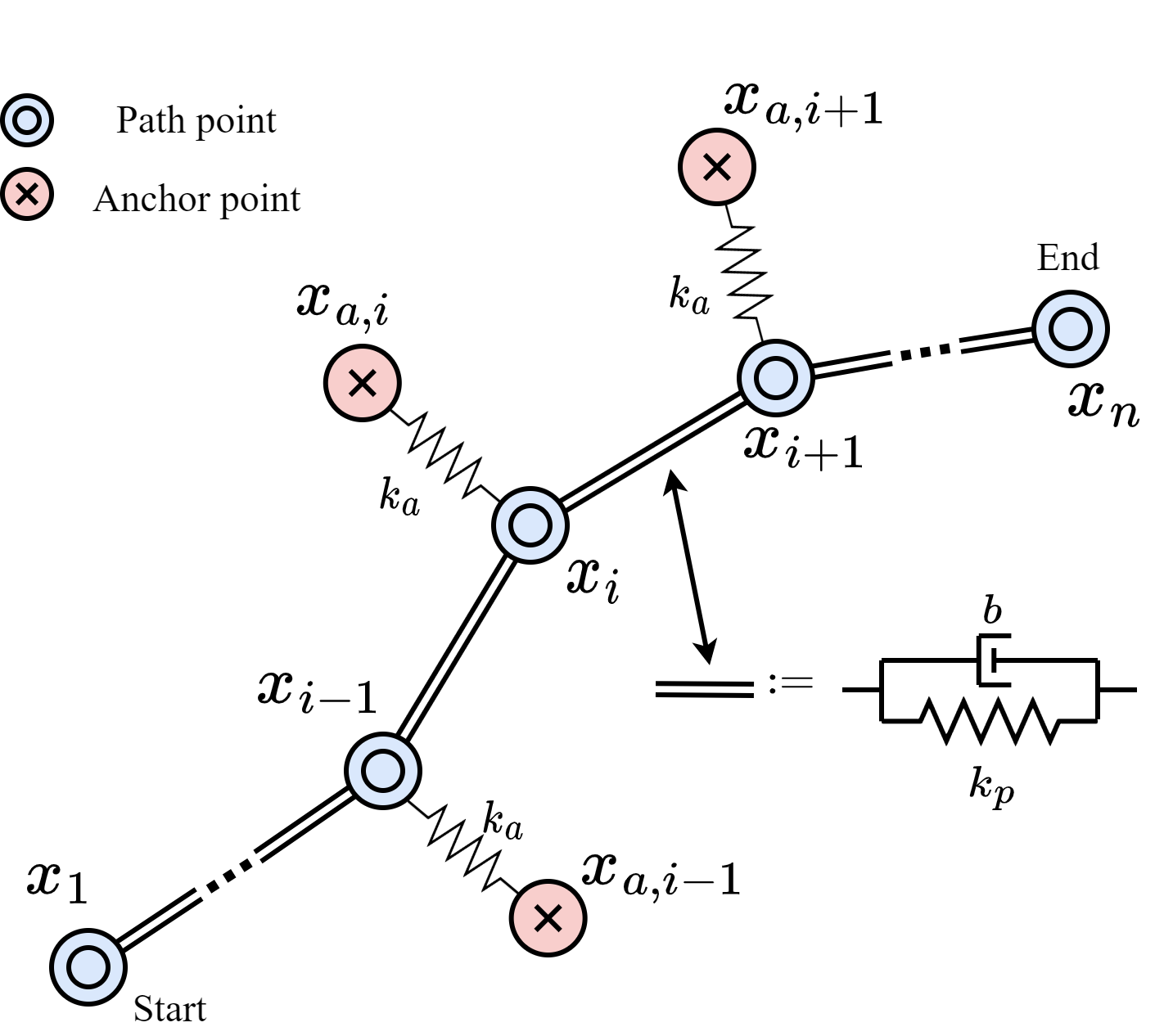}}
    \caption{\textit{\rvp system representation of path points (blue) and anchor points (red). The path points $x_i$ are a discrete representation of the local path and are connected to adjacent path points via a spring and damper while the anchor points $x_{a,i}$ are connected to their corresponding path point via a spring.}}
    \label{fig:system}
    \vspace{-1.70em}
\end{figure}

\emph{Autonomous Navigation} of RVs can be classified into two over-arching categories: \emph{Classical Techniques} and \emph{Learning based techniques} as discussed in the survey paper \cite{xiao2020motion}. 

\emph{Classical techniques} rely on breaking down the problem of \emph{Autonomous Navigation} into 2 main components: (i) \textit{Perception} and (ii) \textit{Navigation}. For perception, RV's use Proprioceptive and Exteroceptive sensors to get a view of the surrounding environment and know the where the RV is located within the environment. \textit{Navigation} can further be split into 3 components: (a) \textit{Path Planning}: Typically translates a higher level abstracted mission to a desired path that the RV needs to take. (b) \textit{Motion Planning} translates this higher level plan into a series of \emph{Motion Primitives} that the RV can then follow. (c) \textit{Dynamic Control} translates the Motion primitives into low level actuator commands such as Acceleration, Steering wheel angle inputs, etc. \emph{Classical techniques} usually rely on a model for the above components and perform a series of optimizations to achieve a goal of \emph{Autonomous Navigation along a desired trajectory avoiding the obstacles}. This work would fall into the category of \emph{Classical techniques}.

\emph{Learning Based techniques} usually employ a data driven model using learning representations to come up with a policy the can replace one of the components (or all of the components such as \textit{Perception, Motion Planning, Path Planning and Dynamic Control}). The advantages of Learning based techniques include  reduced modeling  effort, which oftentimes come with a lack of explainability.

 \cite{508439} introduced Probablistic Roadmap Method (PRM), a sampling based technique. \cite{amato1996randomized} used the PRM technique to create an obstacle based PRM for 3D workspaces, which was one of the first of it's kind to implement the technique for motion planning. \cite{lavalle1998rapidly} introduced Rapidly Exploring Random Trees (RRT) which spawned variants such as RRT* \cite{karaman2011samplingbased}.

\par

Such abstractions can be used with graph based paradigms to obtain Coarse Global Planners like: Dijkstra's shortest path algorithm \cite{dijkstra1959note} and A* heuristic based search algorithm \cite{hart1968formal}. D* lite \cite{koenig2002d} is another optimized version of A* algorithm which avoids re-planning by using smart data structures to increase update speed and save computation cycles.

Local planners usually generate a sequence of finer way-points and/or low-level commands, which are fed directly to the controller to perform a smaller horizon based way-point following. As opposed to the global planner, the local planner needs to take into account the local features surrounding the robot like: obstacles, gradients, etc. The Local Planner is an Online Algorithm that needs to take quick decisions based on the immediate horizon.

Potential Field Methods (PFMs) were first introduced in  \cite{KhattibAPF}. Such an algorithm assumes a potential field that is being generated by the target points such that their gradients change towards the target results in an attractive force towards the target. Such a technique has potential of getting struck in Local Minima that could be generated by the potential fields and results in stagnation of the RV. \cite{ShangguanAPFAgriBot}\cite{weerakoon2015artificial}\cite{lee2012new}\cite{bounini2017modified}\cite{4021735SimAnn} demonstrated optimizations on the existing algorithm to solve the issue of stagnations, but they mostly work under special conditions.
\par 
In the case of well-known vehicle dynamics and the observed environment, \emph{Active Exploration Methods (AEMs)} such as Receding-Horizon Model Predictive Control (MPC) techniques have been used to solve online optimization problems for the purpose of navigation  \cite{Hirose2019MPCVisual}\cite{pan2019agile}\cite{7458179MPC}. Such techniques can be used to generate low level control commands as well as a path for navigation. \cite{Hirose2019MPCVisual} used a MPC controller with a surround view camera to perform navigation and obstacle avoidance in a moving goal fashion. \emph{Partially Observable Markov Decision Processes, (POMDP)} techniques can also be used 
\cite{zhang2017deep}\cite{zhelo2018curiositydriven}. AEMs typically optimize an overall cost with respect to time horizons. AEMs are computationally expensive unless optimizations are done to avoid a large search space. 

\cite{sarvesh2022reshaping} introduced \emph{Reshaping Local Path Planner}, that creates a Local Path based on a \emph{Path Aware Moment Field}. This overcomes the disadvantages of the PFMs and avoids the issue of stagnation and is also computationally more efficient than AEMs by optimizing the action search space.

The classical techniques, albeit successful, require a large amount of engineering effort in modelling and designing the best Local Planner which works for every scenario. Oftentimes, the conventional techniques require extensive modeling which might be difficult to implement. To potentially reduce the amount of engineering effort, data intensive techniques using Machine Learning have been proposed in the literature. There are a large number of publications that address this problem in a few different ways. Based on the survey paper \cite{xiao2020motion}, learning-based navigation techniques can be classified into three parts:
(i) Learning the Entire Navigation Stack, (ii) Learning Navigation Subsystems, (iii) Learning Individual Components. We list a few Learning based planners that are relevant to our research here \cite{pan2019agile}\cite{siva2019robot}\cite{muller2006off}\cite{wigness2018robot}.

Functionally, \rvpfull (RVP) is similar to Elastic Band Planners (EBP) \cite{quinlan1993elastic}. EBP updates and optimizes the Local path incrementally by following the Global path. The elastic band is deformed and it's shape and configuration is changed according to the internal contraction forces along the Global path and external repulsion from the obstacles around the path to remove any slack in the path. In this paper, \rvp is compared with Reshaping Local Path Planner \cite{sarvesh2022reshaping}, which is also a Planner that reshapes a given Global Plan based on Sensor Observations of the Environment.

While other methods such as EBP employ the use of virtual forces that expand and contract the Local path, \rvpfull adds several key elements to produce stable and desirable results. The use of dampers within the Spring-Mass-Damper network provides a more steady reshaping of the path in reaction to (i) \emph{Virtual Obstacle Forces}. Additionally, (ii) \emph{Anchor points} within the system help to retain the shape of the desired Global path and provide key support for path curvature.

\section{Preliminary Definitions}
\subsubsection{Operating Region $\Omega$ }
Consider the operating region for the algorithm as being represented by a two-dimensional region $\Omega$, and having properties defined on it, such as height, surface friction, etc.
\subsubsection{Desired Global Path $\mathbf{x}^{dGP}$ } 
We assume that an upstream \emph{Global Path Planner} has determined a \emph{Desired Global Path} given as a sequence of $n_{GP}$ points $\mathbf{x}^{dGP} = \{ x_i \}, x_i \in \Omega$, $i = \{1, ..., n_{GP}\}$. It is assumed that such a Desired Global Path is generated based on static or stale information. A distance parameterization of the desired global path will also be inferred from $\mathbf{x}^{dGP}$ by calculating the distance along the path: $\mathbf{s}^{dGP} := \{s_i^{dGP} = || x_{i+1}^{dGP} - x_i^{dGP} ||\}  \forall i = \{1, ..., n_{GP} - 1\}, s_1^{dGP} = 0$.
\subsubsection{Sensed Environment $\mathbf{x}^{\epsilon}$ } 
Consider that all the sensors on-board the vehicle have been processed to generate a set of $n_{\epsilon}$ points $\mathbf{x}^{\epsilon} = \{ x_i\}, x_i \in \Omega_{\epsilon} \subset \Omega$, $i = \{1, ..., n_{\epsilon} \}$. Associated with each point is a property function $\epsilon(x): \Omega_{\epsilon} \rightarrow [0, c] $ for some positive constant $c$. In practice this property mapping could be from $\Omega_{\epsilon}$ to integers, or just a boolean, in which case it defines a simple occupancy map. The sensed environment is generated at every control sample, and is considered fresh information. In this paper, the sensed environment is considered to already factor in uncertainties in the sensed information and is presumed to be the best estimate of the environment.

\subsubsection{Local Path $\mathbf{x}^{lP}$ } The output of the primary algorithms of the paper is a sequence of $n_{lP}$ points $\mathbf{x}^{lP} = \{ x_i \}, x_i \in \Omega$, $i = \{1, \hdots, n_{lP} \}$, such that there exists a corresponding smooth curve $\mathbf{\bar{x}}^{lP}(s): [0, s_{lP}] \rightarrow \Omega$, and a monotonic sequence $\mathbf{s}^{lp} = \{ s_i \}, s_i \in [0, s_{lP}]$ such that $\mathbf{\bar{x}}^{lP}(s_i) = \mathbf{x}^{lP}[i]$. Further, this curve shall be sufficiently smooth that a curvature can be defined at every point of the local path, i.e. there exists a well-defined curvature function $\bar{\kappa}^{lP}: [0, s_{lP}] \rightarrow \Re $, and from which we define the sequence $\mathbf{\kappa}^{lP}$ by $\bar{\kappa}^{lP}(s_i) = \mathbf{\kappa}^{lP}[i]$. The path length is $p_{lP} := s_{n_{lP}}$
\subsubsection{Anchor points $\mathbf{x}^a$ } The formulation of the mass-spring-damper system involves a set of anchor points $\mathbf{x}^a = \{x_{a,i} \} \in \Omega$, $i \ \{ 1, \hdots, n_{lP}-2 \}$ corresponding to each of the local path points $x_i \in \mathbf{x}^{lP}$ excluding the start and end points. These anchor points maintain curvature in the path and tether the local path to the global path. Each anchor point $x_{a,i}$ spring has a spring length $l_{a,i}$.
\subsubsection{Spring-Mass-Damper system} The viscoelastic string path $\mathbf{x}^{lP}$ consists of point masses with mass $m$. The spring and damping constants are defined based on two tuning constants $\omega$ and $\zeta$: the spring constant between the path points $k_p \coloneqq m \omega^2$, the spring constant for the anchor points $k_a \coloneqq c k_p$, and the damping constant $b \coloneqq 2m \zeta \omega$. Here, $c$ is a scaling parameter representing the stiffness of $k_a$ relative to $k_p$. The spring lengths $l_{j,k}$ are the intial spring lengths between two points $x_j, x_k \in \Omega$.
\subsubsection{cross track error $e^P$ }
Given a path $\mathbf{x}^P$ and a point $y \in \Omega$, the cross track error $e^P(y)$ is the shortest distance from $y$ to the path, obtained as $||\vec{y y_{\perp}}||$  where the point $y_{\perp}$ is a point on $\mathbf{x}^P$ such that the tangent to the path at that point is $\hat{t}_{y_{\perp}}$, and satisfies $\vec{y y_{\perp}}$ $ \perp \hat{t}_{y_{\perp}}$.

\subsubsection{Path Deviation $\Delta P(\mathbf{x}^1, \mathbf{x}^2)$}
Given two paths $\mathbf{x}^1$ and $\mathbf{x}^2$, the \emph{Path Deviation} is defined as the cumulative sum of the square of the cross track errors: $\Delta P(\mathbf{x}^1, \mathbf{x}^2) := \sum_{x_i} e^2(x_i), \; \; \forall x_i \in \mathbf{x}^1$ 
\subsubsection{Collision $\delta_c (x^P, x^{\epsilon})$}
Given a path $\mathbf{x}^P$ in a sensed environment $\mathbf{x}^{\epsilon}$, we define the boolean collision function $\delta_c$ as true if there is a collision between the path and any of the obstacles, else false. Specifically, we define $\delta_c = (||x^P_i, x^{\epsilon}_j||) > d_c, \; \forall x^P_i \in \mathbf{x}^P, x^{\epsilon}_j \in \mathbf{x}^{\epsilon}$, where $d_c$ is a threshold distance used to determine if the path is too close to obstacles.

\subsection{Problem Statement}
Given a desired global path : $\mathbf{x}^{dGP}$, the problem objective is to synthesize a local path $\mathbf{x}^{lP}$ of length $l_{lP}$ such that there are no collisions, i.e $! \delta_c $, and the Path Deviation $\Delta P (\mathbf{x}^{lP}, \mathbf{x}^{dGP})$ is minimized.
\begin{equation}
    f_{\text{RVP}}: \mathbf{x}^{dGP} \times \Omega \rightarrow \mathbf{x}^{lp}
\end{equation}

\section{Reshaping Viscoelastic-String Planner}

The Local path planner presented here employs the concept of an adaptive viscoelastic path that changes its shape based on applied virtual forces. These virtual forces are a function of the surrounding obstacles and the distance to those obstacles, which is represented by the sensed environment $\mathbf{x}^\epsilon$. For this implementation, the adaptive path is represented by a set of point masses connected in sequence by springs and dampers as shown in Figure \ref{fig:system}. This serves to ensure the path points move as a collective, with adjacent points influencing the movement of one another. Additionally, the path points are anchored to the desired Global path by another set of springs to a set of Anchor points $\mathbf{x}^a$. These Anchor points tether the path points to the original path to ensure the resultant path is similar to the desired Global path and maintains the intended curvature.

The path points $x_i \in \mathbf{x}^{lP}$ are the set of points representing the path within the operating region $\Omega$. The movement of these path points can be represented by the ordinary differential equation (ODE) \eqref{planner:ode}.
\begin{align}
    m\frac{d^2 x_i}{dt^2} + b \frac{d x_i}{dt} = F_{p,i} + \int\limits_r \int\limits_\theta F_{o,i}(r,\theta) \label{planner:ode}
\end{align}
 
There are two virtual forces acting on the path points: a path-based force \emph{$F_{p,i}$} and an obstacle-based force \emph{$F_{o,i}$}.

\subsection{Forces}

\subsubsection{Path Force}
The path-based force $F_{p,i}$ is the force acting on a point $x_i  \in \mathbf{x}^{lP}$ by the adjacent points $x_{i-1}$ and $x_{i+1}$ through the connecting springs in addition to the force from the Anchor points by the point $x_{a,i}  \in \mathbf{x}^a$. This force is calculated according to equation \eqref{planner:path_force}.
\begin{align}
F_{p,i} = k_p &(x_{i-1} - x_i - l_{i,i-1}) + k_p (x_{i+1} - x_i - l_{i,i+1}) \nonumber \\ &+ k_a (x_{a,i} - x_i - l_{a,i}) \label{planner:path_force}
\end{align}
Where $k_p$ is the spring constant between the path points, $k_a$ is the spring constant from the anchor points, and $l$ is the spring length.

\paragraph{Anchor Point Locations}
The Anchor point locations are calculated at the initial step of the simulation by solving for $x_{a,i} \in \mathbf{x}^a$, the location of the anchor point corresponding to the point $x_i \in \mathbf{x}^{lP}$, in equation \eqref{planner:anchor_calc}. This point $x_{a,i}$ creates an equilibrium between the various spring forces. The Anchor points $\mathbf{x}^a$ serve to maintain the path's initial curvature and supply a force for the Local path to retain the shape of the desired Global path provides some incentive for the path to remain unchanged.
\begin{align}
    0 &= k_p (x_{i-1} - x_i - l_{i,i-1}) \nonumber \\ &+ k_p (x_{i+1} - x_i - l_{i,i+1}) + k_a (x_{a,i} - x_i) \label{planner:anchor_calc} \\
    l_{a,i} &= d(x_{a,i}, x_i)
\end{align}

\subsubsection{Obstacle Force}
The obstacle force $F_{o,i}(r,\theta)$ is the force acting on the point $x_i$ from an obstacle at distance $r$ and angle $\theta$. This force is calculated according to the piece-wise function \eqref{planner:obj_force}.
\begin{equation}
    F_{o,i}(r,\theta) = \left\{
    \begin{array}{ll}
        a_2 \cdot \delta_{\mathbf{x}^{\epsilon}}(x_i, r,\theta) \cdot \frac{1}{r^n} & r \leq a_1 \\
        a_2 \cdot \exp{\left( \frac{-(r-a_1)}{a_3} \right)} \cdot \delta_{\mathbf{x}^{\epsilon}}(x_i, r,\theta)  \cdot \frac{1}{r^n} & r > a_1
    \end{array}
    \right .
    \label{planner:obj_force}
\end{equation}
Where within a radius $a_1$, the obstacle force is as defined, and beyond which there is an exponential decay in the force affected by the tuning constant $a_3$. The obstacle force is inversely proportional to the distance $r$ between the path point and obstacle to the power $n$. The force is also scaled by the tuning constant $a_2$. Given a sensed environment $\mathbf{x}^{\epsilon}$, a boolean function $\delta_{\mathbf{x}^{\epsilon}}(x_i, r,\theta)$ is defined to be true if the point of distance $r$ and orientation $\theta$ away from point $x_i$  is occupied in the sensed environment $\mathbf{x}^\epsilon$ and false otherwise. This function ensures that the force only acts on the path point if there is an obstacle at a distance $r$ and angle $\theta$.

\begin{figure}[htbp]
    \vspace{-0.5em}
    \centerline{\includegraphics[width=0.65\columnwidth]{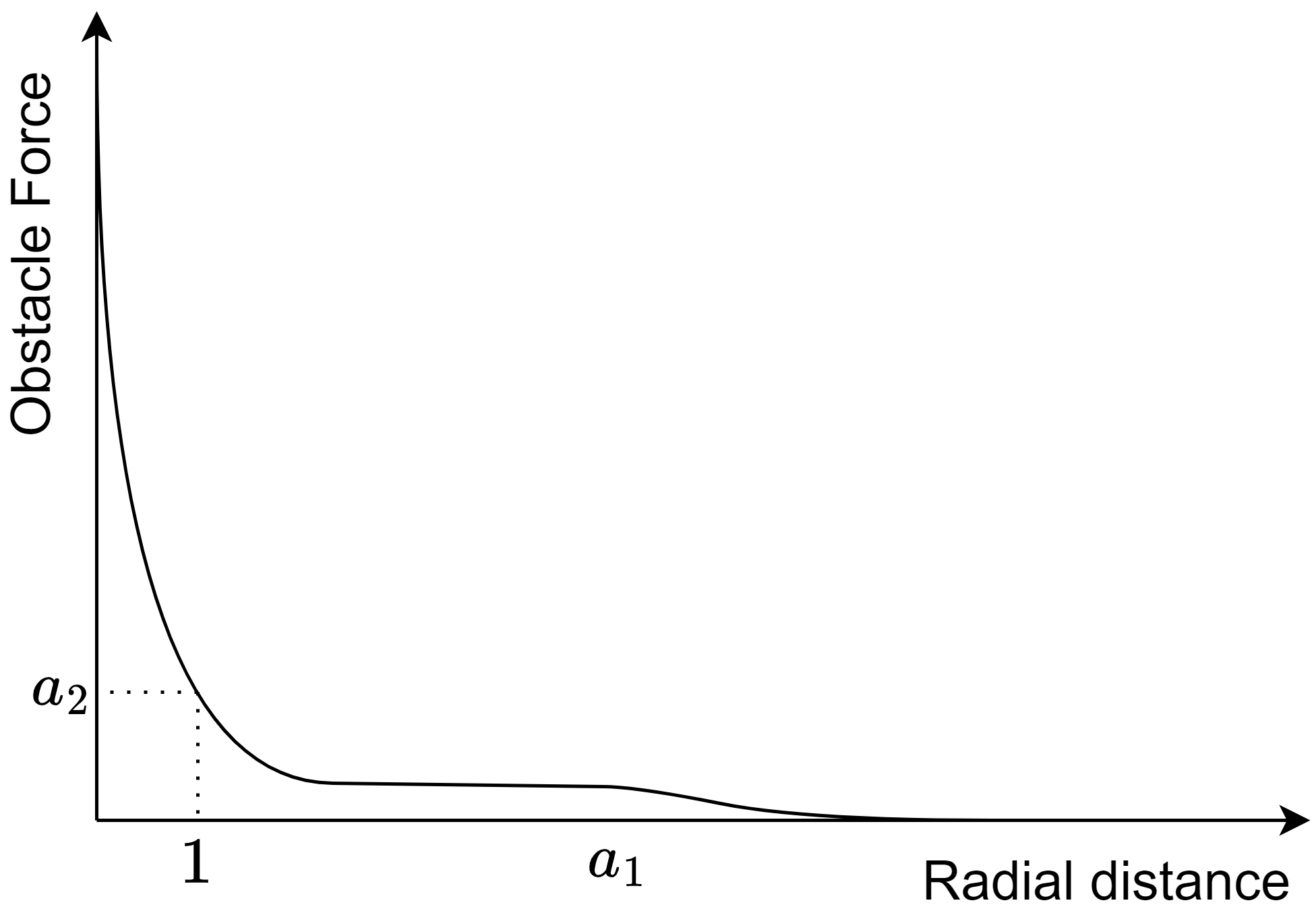}}
    \caption{\textit{Obstacle force as a function of radial distance $r$ for tuning parameters $a_1,a_2$.}}
    \label{fig:obj_force}
\end{figure}

The profile of the obstacle force as a function of the distance $r$ can be seen in Figure \ref{fig:obj_force}.

The force is very high at close distances to the obstacle and rapidly decreases due to the inverse proportionality to distance $r$. After the cutoff distance $a_1$, the force decays and eventually reaches a negligible magnitude. This ensures that obstacles further away from the path point enact little to no force. 

\subsection{Resultant Path}

\rvp approximates the motion of the path points until an integration time $t_f$, which can also be represented by a maximum number of steps as in Algorithm \ref{alg:planner}. If the system detects that the path points $\mathbf{x}^{lP}$ have reached a steady-state prior to reaching $t_f$, it will automatically exit the simulation to decrease computational expense. The simulation considers the steady state to be satisfied if all path points have an acceleration below a certain threshold $a_t$: $\forall x \in \mathbf{x}^{lP}, \Ddot{x} < a_t$.

After reaching the final state of the path, a spline-fit is performed on the path points and is interpolated to produce a set of evenly spaced points, which is the output of the \rvpfull.

\begin{figure}[hbtp]
    \centering
    \subfigure[\textit{No obstacle proximity}]
    {
        \centering
        \includegraphics[width=0.22\textwidth]{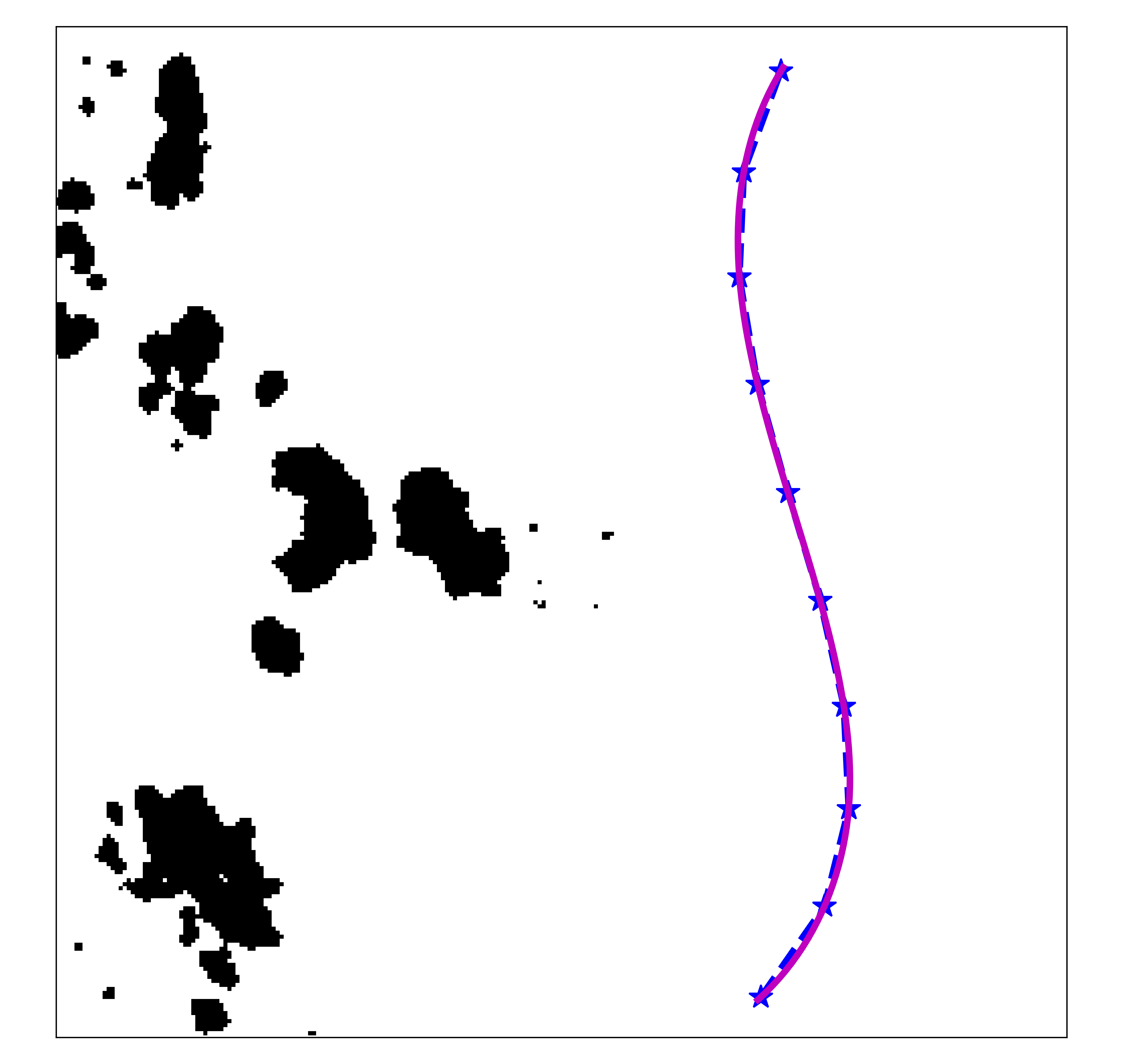}
        \label{fig:planner_performance:none}
    }
    \subfigure[\textit{Simple reshaping}]
    {
        \centering
        \includegraphics[width=0.22\textwidth]{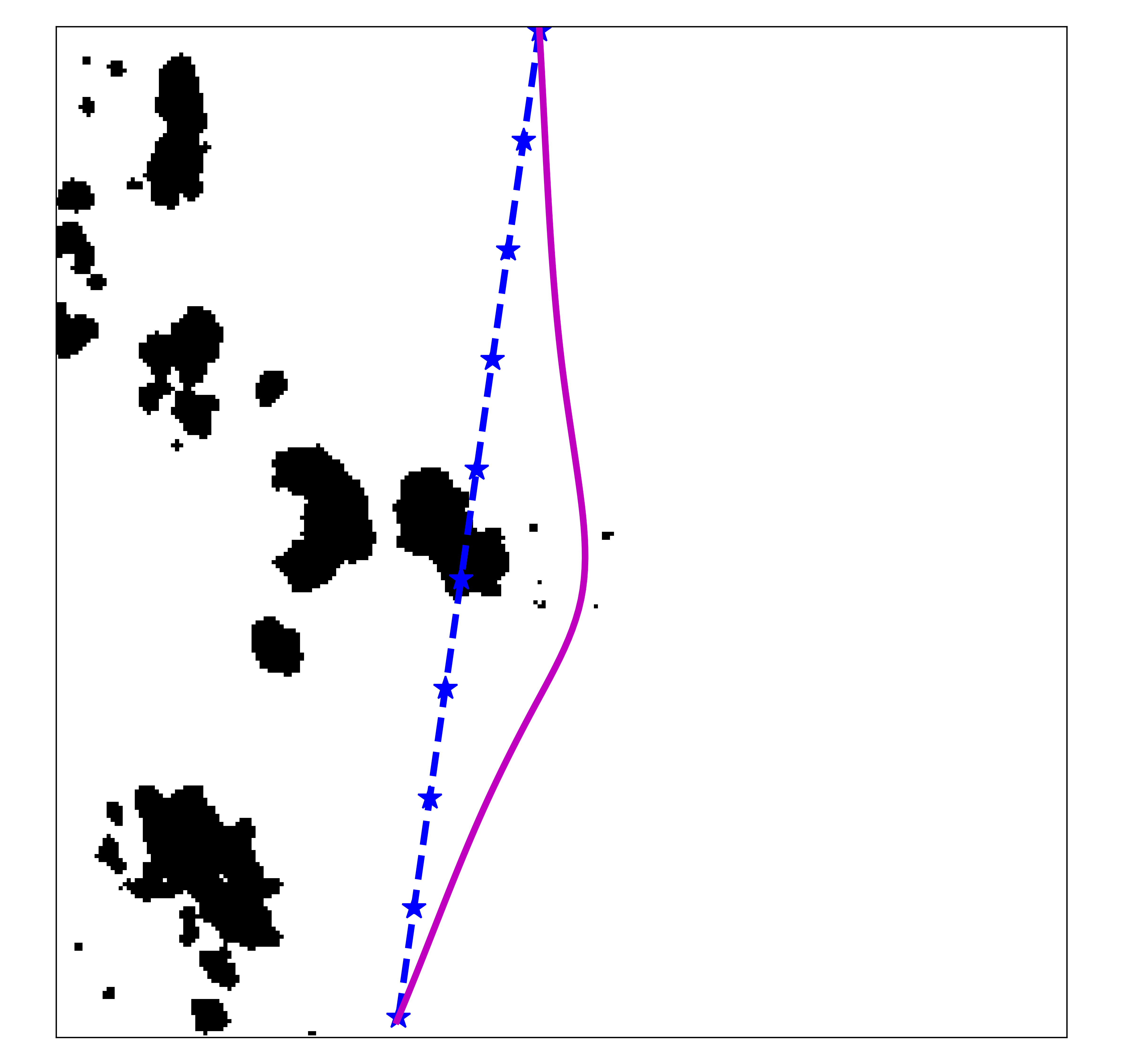}
        \label{fig:planner_performance:simple}
    }
    
    \subfigure[\textit{Complex reshaping}]
    {
        \centering
        \includegraphics[width=0.22\textwidth]{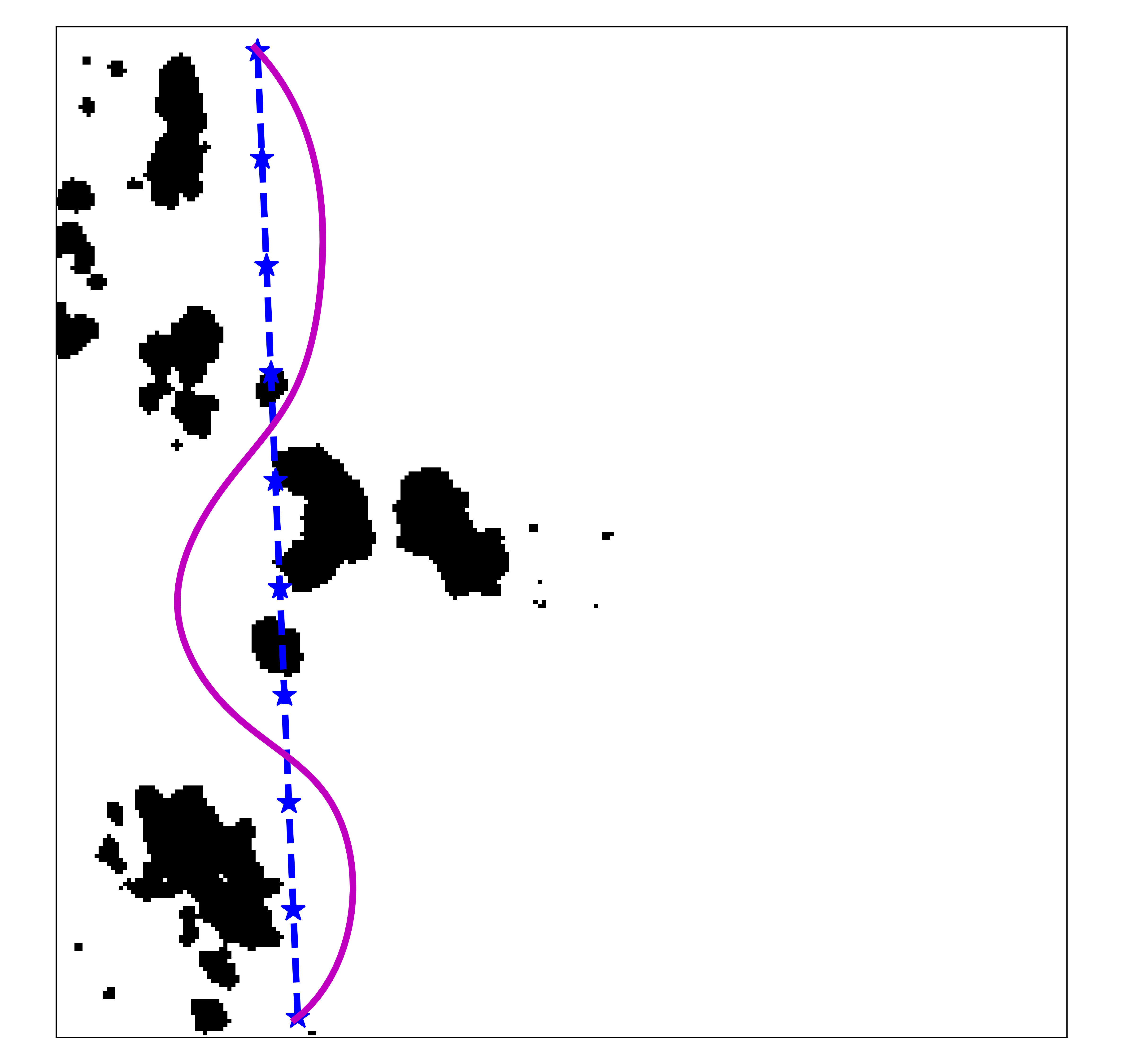}
        \label{fig:planner_performance:complex}
    }
    \subfigure[\textit{Complex obstacle shape}]
    {
        \centering
        \includegraphics[width=0.22\textwidth]{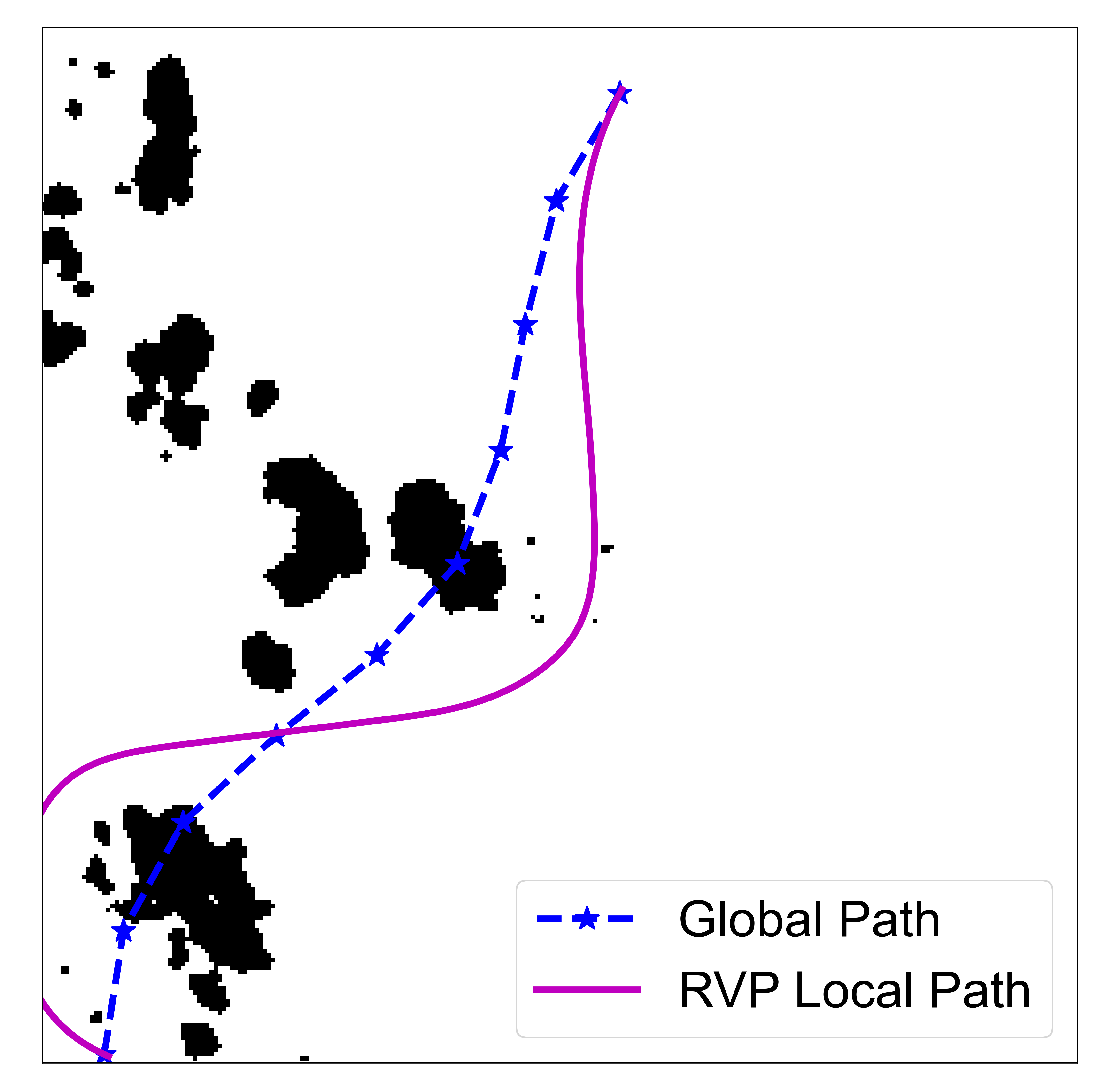}
        \label{fig:planner_performance:complex2}
    }
    \caption{\textit{RVP performance over four scenarios over data from a Real-Unstructured Environment. The Global path is shown in dashed blue, the reshaped Local path is shown in solid magenta. (a) demonstrates the output of \rvp when the desired global path is not in close proximity to any obstacles. (b) demonstrates the output of \rvp when there is a single obstacle obstructing the desired global path. (c) \& (d) demonstrate increasingly complex scenarios where multiple obstacles overlap the desired global path and there is a significant reshaping effort needed to avoid the obstacles.}}
    \label{fig:planner_performance}
\end{figure}

The path planner is primarily evaluated based on its ability to provide a safe and stable path in a wide range of scenarios. In Figure \ref{fig:planner_performance}, the \rvpfull is evaluated over four different global paths along with a \emph{real Unstructured Sensed environment based on real Lidar data} obtained in an abandoned Quarry at the RELLIS campus of Texas A\&M University. These select examples with varying difficulty highlight the large range of scenarios the planner can handle. 

In Figure \ref{fig:planner_performance:none}, the global path is not in close proximity to any obstacles. The resulting reshaped path from the planner therefore does not experience any forces from the obstacles and maintains the original path between due to the anchor points keeping the system in equilibrium. In Figure \ref{fig:planner_performance:simple}, there is one obstacle obstructing the global path, which causes the path to be pushed outward. However, the path forces ensure that the provided reshaped path does not deviate very far from the global path. In Figure \ref{fig:planner_performance:complex}, the global path must be reshaped significantly to pass through two key areas. The path points are able to reach a steady state passing in between the obstacles after an equilibrium is found between obstacle and path forces. Figure \ref{fig:planner_performance:complex2} indicates another example where the global path must undergo notable changes, in particular due to the oddly-shaped obstacle in the bottom left. The planner provides a path that avoids all obstacles and handles the complex obstacles.

\begin{figure}[htbp]
    \centerline{\includegraphics[width=1.0\columnwidth]{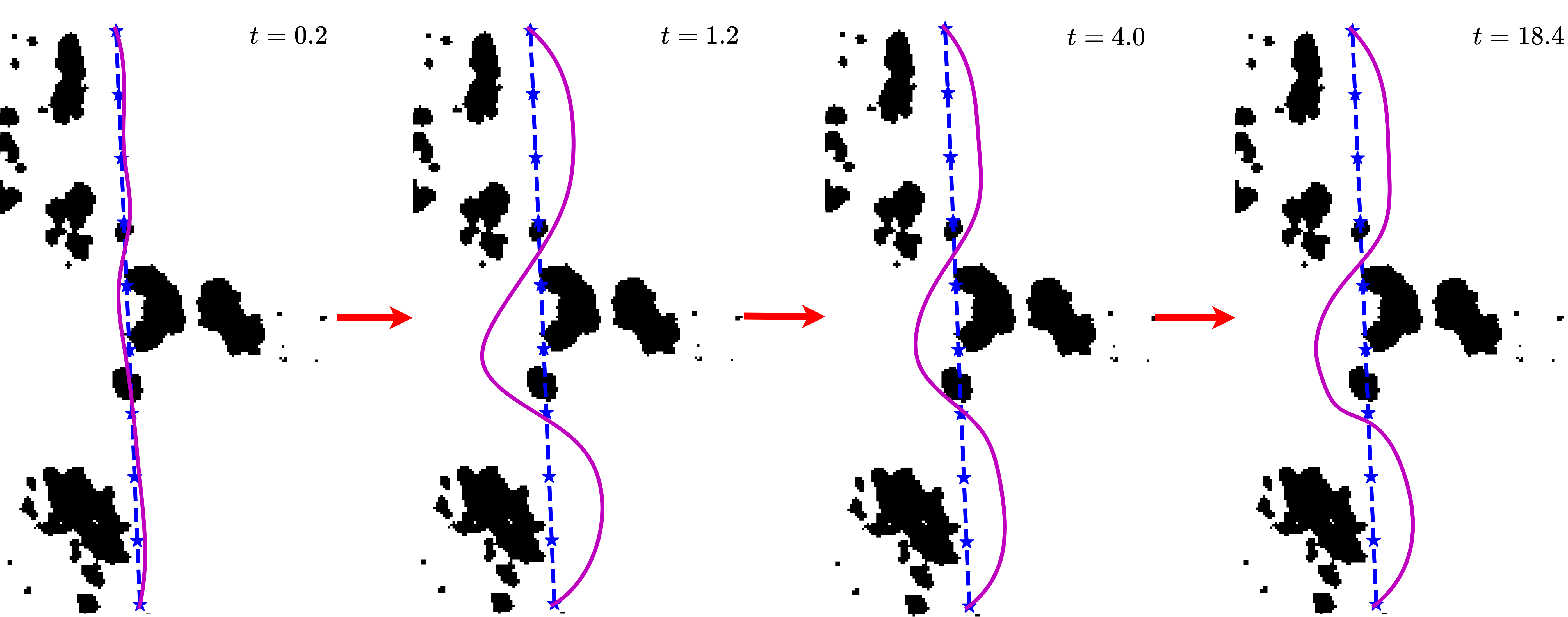}}
    \caption{\textit{Visualization of dynamically reshaped Local path at various integration timesteps for scenario in Figure \ref{fig:planner_performance:complex}. The Global path is shown in dashed blue and the current state of the Local path in solid magenta.}}
    \label{fig:breakdown}
\end{figure}

In Figure \ref{fig:breakdown}, the reshaped Local path is shown at select integration times $0 < t < t_f$ during the reshaping process. At the start, the path points $x_i \in \mathbf{x}^{lP}$ are initially being pushed outward by the obstacle forces. In the second integration timestep, the path points reach their peak deviation after which the path forces pull the path points back towards the initial path in the third integration timestep. After some refinement, the Local path finally reaches a steady state unobstructed by obstacles due to the damping of the system. This dynamic movement of the Local path indicates how the presented path planner is robust to a variety of paths and sensed environments.

\subsection{Planner Flow}

Algorithm \ref{alg:planner} provides a high level overview of the inner workings of this planner when finding a solution using discrete methods. In each step of simulating the path point movement, the path forces and obstacle forces are calculated and used in conjunction with dynamics to move the simulation forward.

\begin{algorithm}[H]
    \caption{\rvpfull}\label{alg:planner}
    \KwData{$\mathbf{x}^{dGP}$: desired global path; $\mathbf{x}^\epsilon$: sensed environment; $\omega, \zeta$: system parameters; $a_1, a_2$: obstacle force parameters, $p_{min}$: minimum portion of steps to simulate}
    \KwResult{$\mathbf{x}^{lP}$: local path}

    $\mathbf{x}^{lP} \leftarrow \mathbf{x}^{dGP}$
    
    $k_p, k_a, b \leftarrow f(\omega, \zeta)$
    
    Initialize anchor points, spring lengths

    $\text{step} \leftarrow 1$
    
    \While{$\text{step} < \text{max\_steps}$}{
        $F_{p} \leftarrow \text{Path forces } \forall x_i \in \mathbf{x}^{lP}$

        $F_{o} \leftarrow \text{Obstacle forces } \forall x_i \in \mathbf{x}^{lP}$

        $\mathbf{x}^{lP} \leftarrow \text{step\_dynamics}(\mathbf{x}^{lP}, F_p, F_o)$

        \If{$\text{path\_stagnated}(\mathbf{x}^{lP})$}{
            perturb\_path$(\mathbf{x}^{lP})$ \Comment{Perturb stagnated path point}
        }

        \If{$\text{step} > p_{min} \cdot \text{max\_steps} \text{ and is\_steady}(\mathbf{x}^{lP})$}{
            \textbf{return} $\mathbf{x}^{lP}$ \Comment{Path has reached steady state}
        }
    }
    \textbf{return} $\mathbf{x}^{lP}$
\end{algorithm}

The planner also addresses common problems encountered by similar Potential Field Methods such as getting stuck in Local Minima, which could result in the path getting stuck in undesirable locations. The functions \emph{path\_stagnated} and \emph{perturb\_path} identify path points that are stuck in local minima inside of obstacles (indicating a collision) and perturb them. This perturbation, in addition to the path forces from adjacent points, is able to pull the stagnated points into more favorable positions.

Additionally, the system simulates a minimum number of steps based on a user-defined variable $p_{min}$. If the system indicates that the path has reached a stable, steady-state solution after those minimum number of steps, no further simulation is needed and the path is returned. This is done in order to save computation time in the case of relatively simple reshaping.

\subsection{Iterative Simulation}

The path does not always reach a safe state after a single iteration of the simulation, so multiple iterations may be required. After each iteration, the path is evaluated for collisions. If the evaluation deems the path to be safe, the new path is returned. Otherwise, the unsafe path is used as the initial path for another iteration to obtain a safer path. Each successive iteration enforces a decay in the system parameters to ensure that later iterations do not drastically alter the changes from the previous iterations. Due to the decay in the algorithm’s parameters, if more than a specified number of iterations are required to reach a safe path, the simulation is exited and the path is flagged as unsafe.

\begin{algorithm}[H]
    \caption{Iterative \rvp}\label{alg:iterative_sim}
    \KwData{$\mathbf{x}^{dGP}$: desired global path; $\mathbf{x}^\epsilon$: sensed environment; $\omega, \zeta$: system parameters; $a_1, a_2$: obstacle force parameters; $\lambda$: parameter decay; $d_c$: safety margin}
    \KwResult{$\mathbf{x}^{lP}$: local path; flag: (1: safe, 0: unsafe)}

    $\mathbf{x}^{lP} \leftarrow \mathbf{x}^{dGP}$
    
    $\text{iteration} \leftarrow 1$
    
    \While{$\delta_c(\mathbf{x}^{lP}, \mathbf{x}^\epsilon, d_c)$}{
        $a_1 \leftarrow a_1 \cdot \lambda^{\text{iteration}-1}$
        
        $a_2 \leftarrow a_2 \cdot \lambda^{\text{iteration}-1}$
        
        $\mathbf{x}^{lP} \leftarrow$ RVP($\mathbf{x}^{lP}$, $\mathbf{x}^\epsilon$, $\omega$, $\zeta$, $a_1$, $a_2$)
        
        \If{$\text{iteration} = \text{max\_iters}$}{
            \textbf{return} $\mathbf{x}^{lP}$, $0$ \Comment{Local path marked as unsafe}
        }
        
        $\text{iteration} \leftarrow \text{iteration} + 1$
    }
    \textbf{return} $\mathbf{x}^{lP}$, $1$ \Comment{Local path is collision free}
\end{algorithm}
\begin{figure}[hbtp]
    \vspace{-1.7em}
    \centering
    \subfigure[\textit{Iteration 1}]
    {
        \centering
        \includegraphics[width=0.19\textwidth]{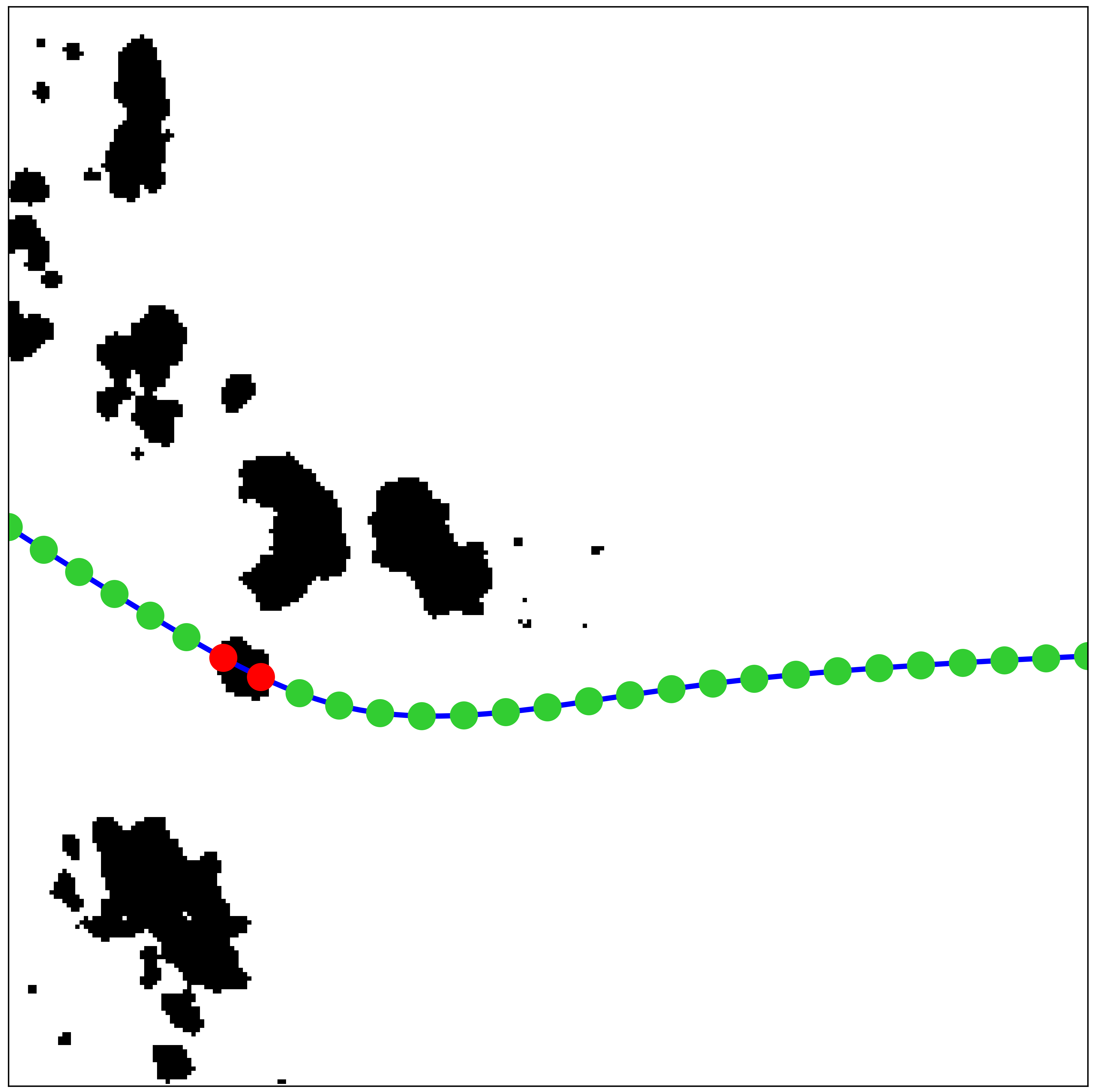}
    }
    \subfigure[\textit{Iteration 2}]
    {
        \centering
        \includegraphics[width=0.19\textwidth]{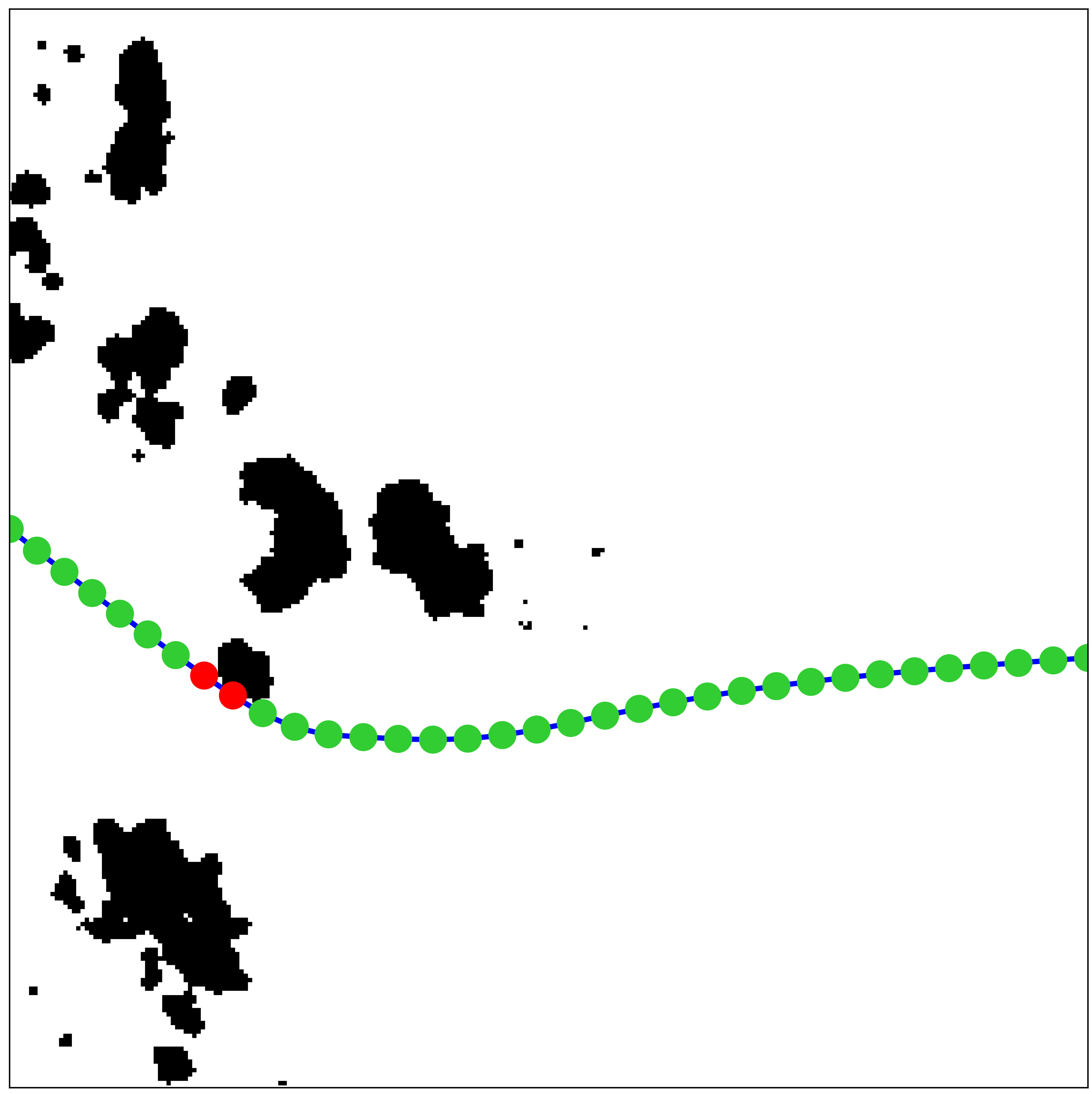}
    }
    
    \subfigure[\textit{Iteration 3}]
    {
        \centering
        \includegraphics[width=0.19\textwidth]{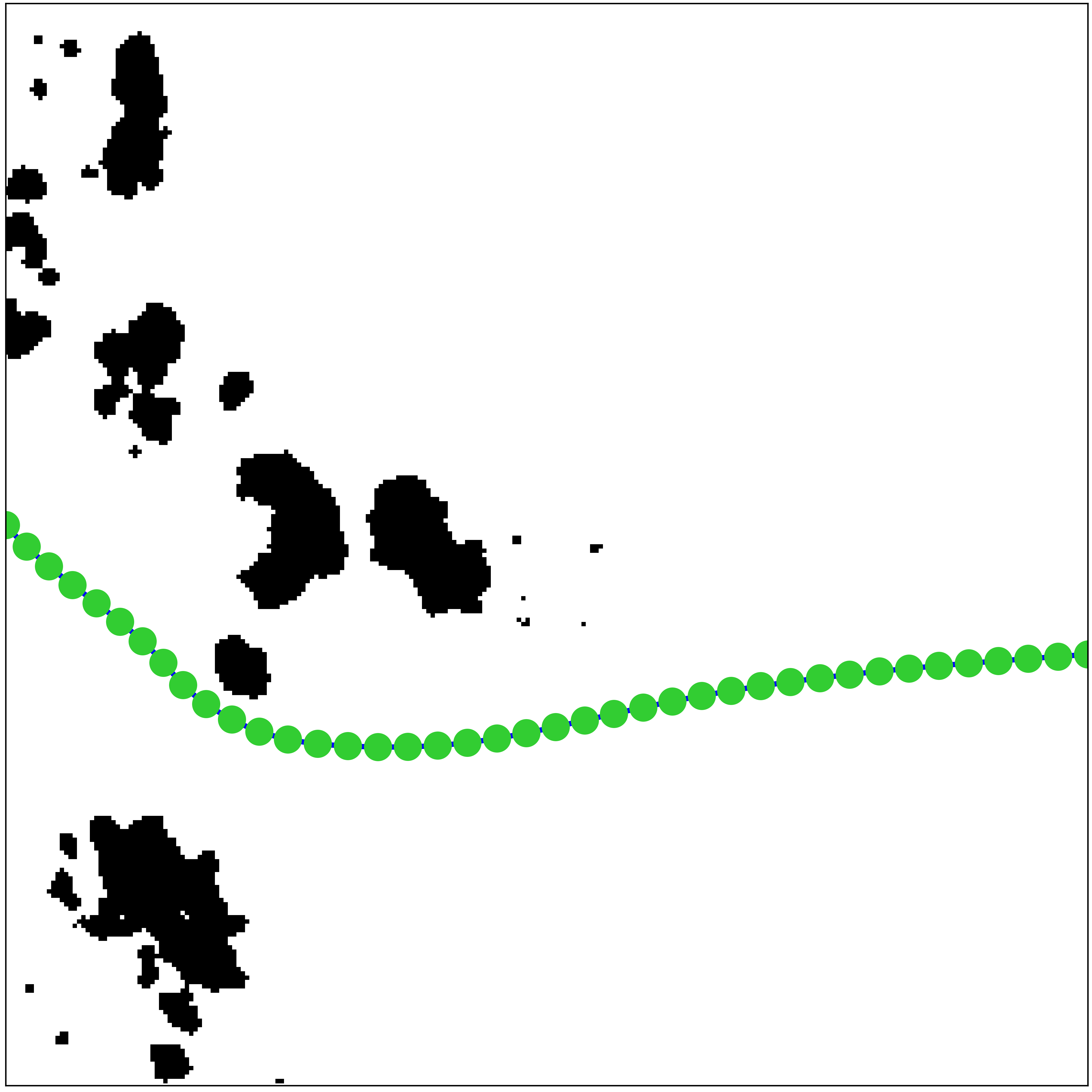}
    }
    \caption{\textit{Progression of model evaluation for a given path and $d_c$. The sensed environment $\mathbf{x}^\epsilon$ is represented by the obstacles in black. The evaluated points are marked green if deemed safe with no collisions $(!\delta_c)$ and marked red if unsafe $(\delta_c)$.}}
    \label{fig:path_eval}
\end{figure}

\subsubsection{Path Evaluation}

The resultant path is evaluated at each iteration for collisions using the boolean collision function $\delta_c(\mathbf{x}^{lP}, \mathbf{x}^\epsilon, d_c)$, which is a function of the sensed environment $\mathbf{x}^\epsilon$ and threshold safety distance $d_c$.

The collision function $\delta_c$ is evaluated $\forall x_i \in \mathbf{x}^{lP}$ and additional points that are interpolated between the points $x_i$, $\mathbf{x}^{\text{itp}}$, which forms the set $\mathbf{x}^{\text{eval}} \coloneqq \mathbf{x}^{lP} \cup \mathbf{x}^{\text{itp}}$. If the path at the end of an iteration is not deemed safe $(\delta_c = \text{true})$, then the set of points $\mathbf{x}_{\text{add}} \coloneqq \{x_i^e \big| \forall x_i^e \in \mathbf{x}^{\text{eval}}, x_j^\epsilon \in \mathbf{x}^\epsilon, || x_i^e, x_j^\epsilon || > d_c \}$ is inserted into the path for the next iteration of the simulation, $\mathbf{x}^{lP}_{\text{iter}=n+1} \coloneqq \mathbf{x}^{lP}_{\text{iter}=n} \cup \mathbf{x}_{\text{add}}$, which is used as the initial path of the next iteration.

Figure \ref{fig:path_eval} shows this iterative simulation in action, with the path eventually reaching a stable safe state, passing all safety requirements after several iterations. The evaluations points are displayed and colored by their measured safety based on a set $d_c$ value.

\section{Results}

\subsection{Planner Validation}

In order to demonstrate the capability of RVP, it is validated in simulation on randomly generated sensed environments and global paths. A large dataset of these scenarios was generated, covering a wide range of possible path-obstacle configurations with varying difficulty.

\subsubsection{Dataset}
\begin{figure}[htbp]
    \centerline{\includegraphics[width=0.6\columnwidth]{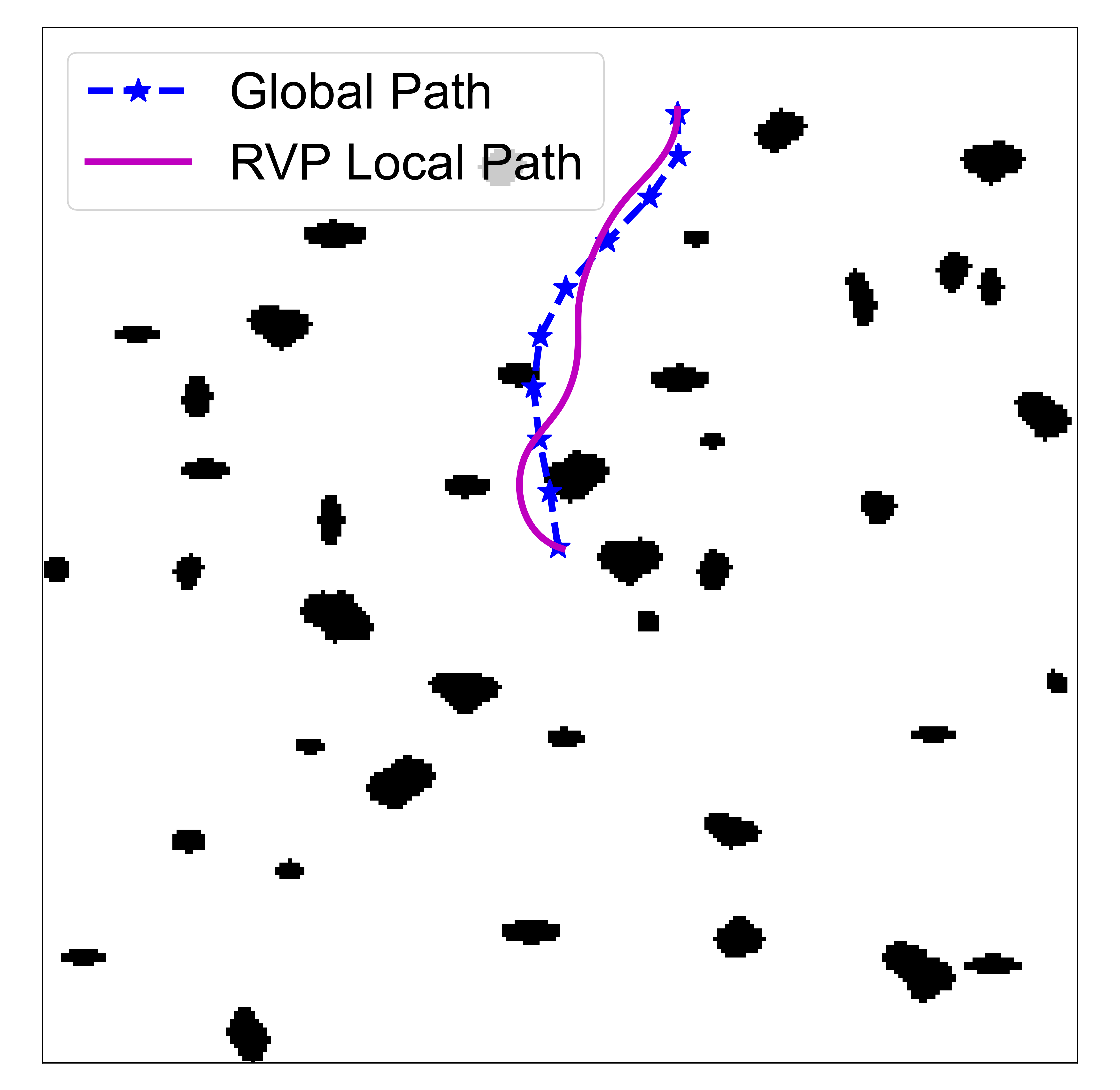}}
    \caption{\textit{An example randomly generated sensed environment and desired Global path with its corresponding RVP reshaped Local path.}}
    \label{fig:rand_scenario}
\end{figure}

Generating a random scenario begins with a random sensed environment. The sensed environment starts as an empty occupancy grid, and a random number of objects are successively placed into the map. These objects have a range of shapes and sizes and are representative of commonly occurring obstacles such as bushes and trees in an Unstructured Environment.
Once the sensed environment is created, a random global path is generated. The distribution of random paths covers a wide range of scenarios, including straight paths between the start and end points as well as extremely curved paths that may intersect objects. 
The diversity of sensed environments and paths in the dataset allows for rigorous testing of the planner. One example scenario is shown in Figure \ref{fig:rand_scenario} with a large number of objects and a random desired Global path.

\subsubsection{Performance}
Evaluating \rvpfull on a dataset of $10,000$ scenarios, we find that it has a success rate of $\mathbf{94.3\%}$. Here, success rate is defined as the local path $\mathbf{x}^{lP}$ having no collisions along the entire path given the sensed environment $\mathbf{x}^\epsilon$ i.e $!\delta_c(\mathbf{x}^{lP}, \mathbf{x}^\epsilon)$.

\subsection{Comparison Scenarios between \rvpfull (\rvp) and Reshaping Local Path Planner (RLP)}
\begin{figure}[hbtp]
    \centering
    \subfigure[]
    {
        \centering
        \includegraphics[width=0.9\columnwidth]{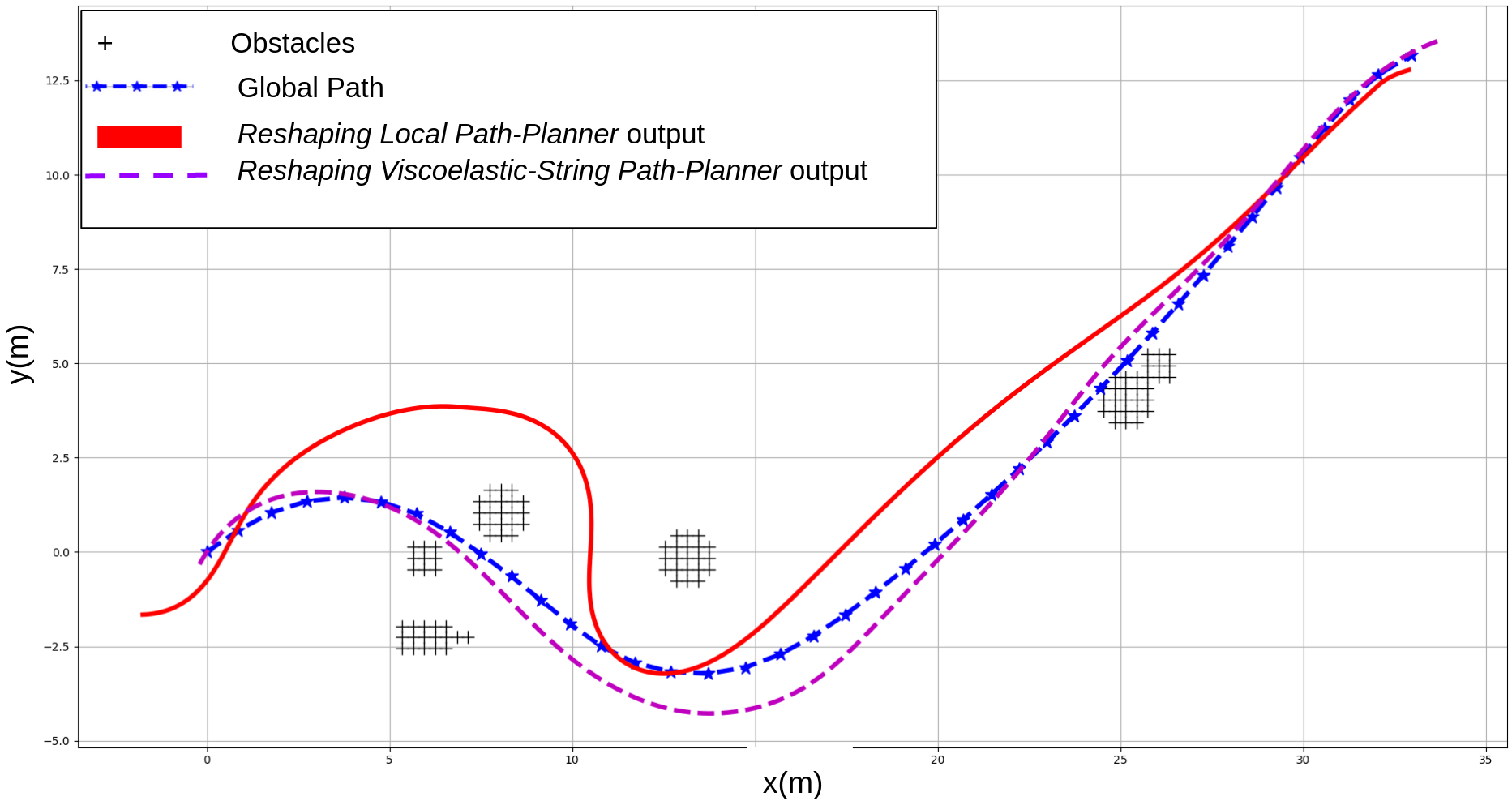}
        \label{fig:ComparisonScen1}
    }
    
    \subfigure[]
    {
        \centering
        \includegraphics[width=0.9\columnwidth]{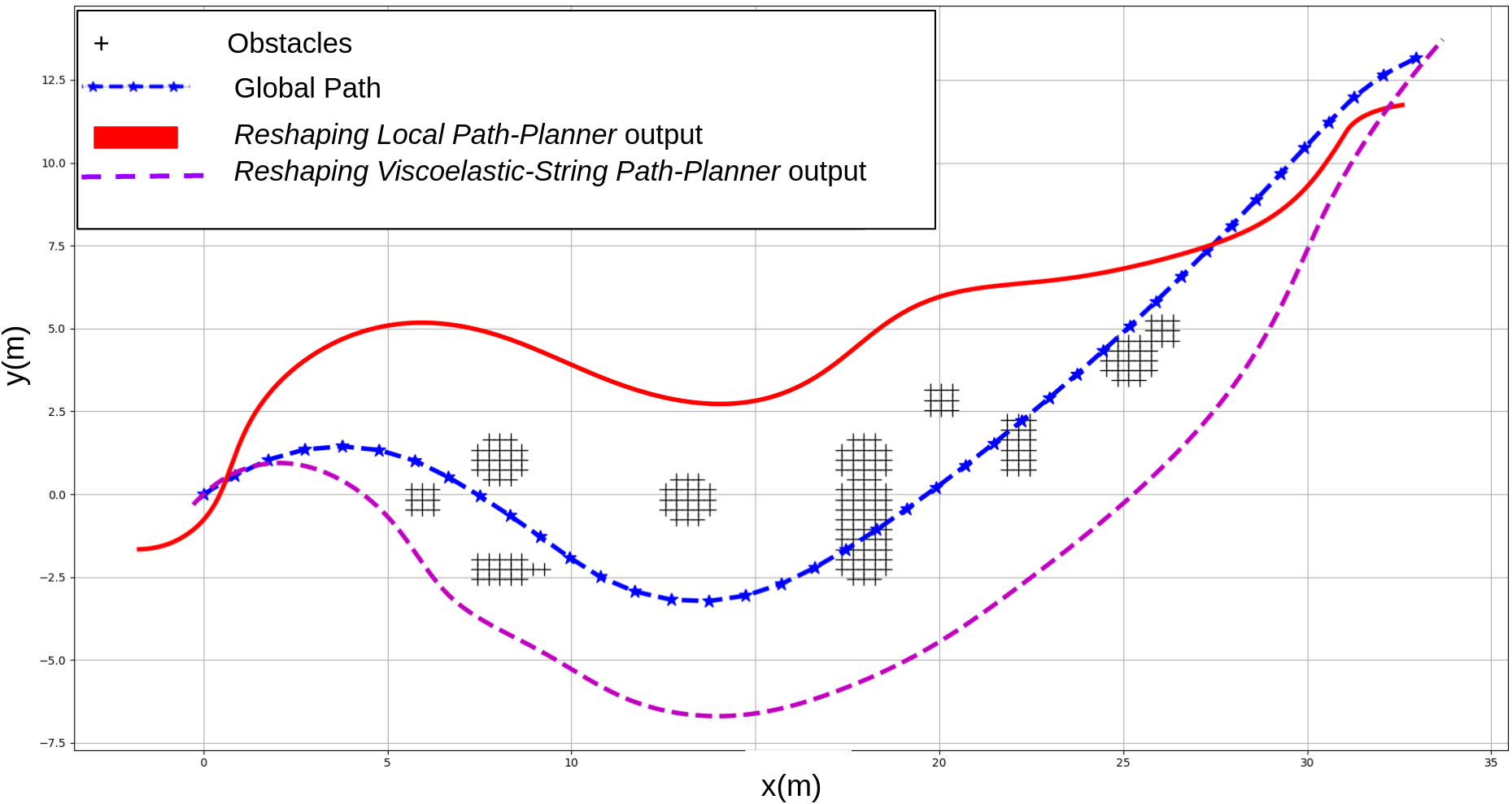}
        \label{fig:ComparisonScen2}
    }
    
    \subfigure[]
    {
        \centering
        \includegraphics[width=0.9\columnwidth]{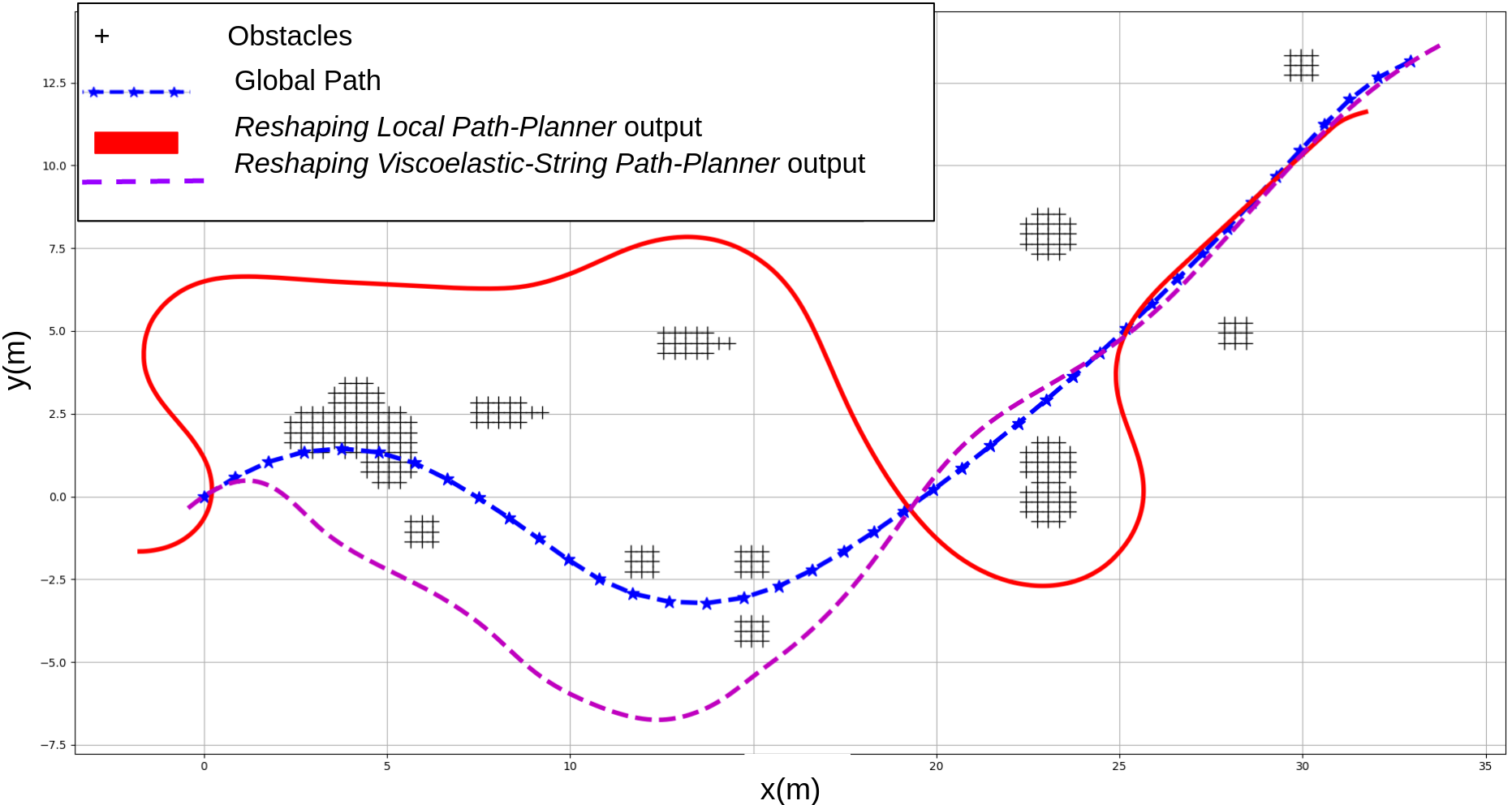}
        \label{fig:ComparisonScen3}
    }
    \caption{\textit{Comparison between Reshaping Local Path Planner and \rvpfull on 3 unstructured environments over a non-straight desired global path.}}
    \label{fig:ComparisonScen}
\end{figure}

We demonstrate the working of the \rvp with a similar existing Local Planner \cite{sarvesh2022reshaping} i.e \emph{Reshaping Local Path Planner (RLP)} in Figure \ref{fig:ComparisonScen}. \cite{sarvesh2022reshaping} also reshapes a given Global Plan in presence of Obstacles using a \emph{Path Agnostic Turning Moment}. In the case of the RLP, it can be seen that the RV starts at a point away from the starting point of the Global Path. Both RLP and \rvp have a look-ahead that captures the entire length of the desired Global Path.

In Figures \ref{fig:ComparisonScen1} and \ref{fig:ComparisonScen3}, \rvp performs clearly better than RLP in maintaining a small path deviation $\Delta P(\mathbf{x}^{dGP}, \mathbf{x}^{lP})$ from the desired global path. In Figure \ref{fig:ComparisonScen2}, the two planners have fairly comparable performance due to the specific path-obstacle configuration. More generally, \rvp performs at least as well as RLP due to the entire path being reshaped simultaneously and the stability built into the system.

\section{Conclusions and Future Work}

This paper introduces a novel reshaping path planner that reshapes a Desired Global Path  based on two key concepts: (i) Virtual obstacle forces that act on the path point masses and (ii) Spring-Mass-Damper system and Anchor Points, which maintain a smooth curvature and steadiness in the resultant local path.

The results of \rvp are also validated on a large randomly generated dataset as well as qualitatively against a similar reshaping planner.

The key drawback of the \rvpfull is the computational expensiveness of approximating the ODE in equation \eqref{planner:ode}. To remedy this issue, we propose training a Neural Network on a dataset of scenarios solved by \rvp in order to have real-time performance. Some initial progress has been made in using a Soft Actor Critic model \cite{haarnoja2018soft} and a Proximal Policy Optimization model \cite{schulman2017proximal} with a specialized reward function to learn the ODE.

\bibliographystyle{IEEEtran}
\bibliography{IEEEabrv,References/conference_101719.bib}

\end{document}